\documentclass{article}
\usepackage{ijcai19}

\usepackage{times}
\usepackage{soul}
\usepackage{url}
\usepackage[hidelinks]{hyperref}
\usepackage[utf8]{inputenc}
\usepackage[small]{caption}
\usepackage{graphicx}
\usepackage{amsmath}
\usepackage{booktabs}
\usepackage{algorithm}
\usepackage{algorithmic}
\usepackage{amssymb,amsfonts,bm,amsthm}
\usepackage{multirow,makecell,footmisc}
\usepackage{enumerate}
\usepackage{lineno}
\usepackage{textcomp}
\usepackage{subfigure}
\usepackage{mathtools}
\usepackage{threeparttable}
\usepackage{color}
\usepackage{appendix}
\usepackage{comment}
\usepackage{blindtext}
\title{Structured Semantic Model supported Deep Neural Network for Click-Through Rate Prediction}

\author{
Chenglei Niu$^1$
\and
Guojing Zhong$^{1}$\and
Ying Liu$^1$\and
Yandong Zhang$^1$\and
Yongsheng Sun$^1$\and
Ailong He$^2$\and
Zhaoji Chen$^1$
\affiliations
$^1$Weibo\\
$^2$Alibaba Group
\emails
\{chenglei3, guojing7, liuying15, yandong7, yongsheng, zhaoji1\}@staff.weibo.com,
along.hal@alibaba-inc.com
}

\begin{document}

%\keywords{Click-Through Rate Prediction, Display Advertising, Recommender System}
\maketitle

\begin{abstract}
With the rapid development of online advertising and recommendation systems, click-through rate prediction is expected to play an increasingly important role.
Recently many DNN-based models which follow a similar Embedding\&Multi-Layer Perceptron (MLP) paradigm, and have achieved good result in image/voice and nlp fields.
Applying Embedding\&MLP in click-through rate prediction is popularized by the Wide\&Deep announced by Google.
%CNN-based models involuve convolution of feature sequences into DNN which could learn deep interaction between features espicially for abundant word segment features, and the number of trainable variables normally grow dramatically as the number of feature fields, the number of convolutions and the sequences dimension grow. It is a big challenge to get state-of-the-art result through training deep neural network and embedding together, since it falls into local optimal or overfitting easily.
These models first map large scale sparse input features into low-dimensional vectors which are transformed to fixed-length vectors, then concatenated together before being fed into MLP to learn the non-linear relations among input features. 
The number of trainable variables normally grow dramatically as the number of feature fields and the embedding dimension grow. 
It is a big challenge to get state-of-the-art result through training deep neural network and embedding together, since it falls into local optimal or overfitting easily.
In this paper, we propose an Structured Semantic Model (SSM) to tackle this challenge by designing an orthogonal base convolution and pooling model which adaptively learns the multi-scale base semantic representation between features supervised by the click label.
The outputs of SSM are then used in the Wide\&Deep for CTR prediction.
Experiments on two public datasets as well as real Weibo production dataset with over 1 billion samples have demonstrated the effectiveness of our proposed approach with superior performance comparing to state-of-the-art methods.
\end{abstract}

%auto-ignore

\section{Introduction}

%https://deepctr.readthedocs.io/en/latest/models/DeepModels.html
Click-through rate (CTR) is widely used in advertising and recommender systems to describe users' preferences on items. For cost-per-click(CPC) advertising system, the revenue is determined by bid price and CTR. For recommender system, CTR is used to improve user experience. So it's important to improve the performance of CTR prediction, and CTR prediction has received much attention from both academia and industry communities.

In recent years, DNN has been researched and applied with structured data for CTR prediction after gaining popularity of deep learning in image, voice, NLP, etc. fields.

Structured data is particularly abundant(such as user behavior data, blog segmentation, user interest, etc.) for CTR prediction, which requires a lot of feature engineering effort with traditional methods.
DNN-based models can take advantages of those structured data and deep feature representation could be learned through embedding and multi-layer perceptrons. 
But it requires training large number of variables and using lots of training epochs which aggravates the risk of overfitting and dramatically increases the computation and storage cost, which might not be tolerated for an industrial online system.

%DNN-based methods could not use sparse categorical features directly due to their high dimensionality nature. 
%The sparse categorical features need to be transformed to low-dimensional dense features.
%The principle of most such transformation methods is Embedding. Research of DNN-based CTR prediction mainly focuses on the design of the embedding layer \cite{embdeep}, such as FNN\cite{FNN}, PNN\cite{PNN}, etc.
There are many researches based on CNN, but due to the particularity of advertising/recommended data, there are no context data like images and voices. 
Therefore, the convolution and pooling over advertisement/recommended data is usually unexplainable.

In the field of search, Microsoft's proposed DSSM\cite{DSSM} model establish as supervised semantic model for doc and query, which learns the non-linear relations of word granularity well.
Inspired by this, we propose an structured semantic model (SSM) to learn semantic representation over features.% supervised by semantic over features together with the rest of DNN variables. 
Additionally, the SSM we propose uses a series of base convolutions instead of traditional trainable convolutions, followed a hidden layer to train convolution variables and poolings together. We named this "delay convolution".
This method effectively learns the multi-scale base semantic representation of the user-item and used in later DNN training.

The contributions of this paper are summarized as follows:
\begin{itemize}
\item We propose a novel approach to train the embedding of sparse feature which effective improves the convergence of DNN-based models than traditional CNN-based models.
The semantic relations between user and item features can be figured out with the structured semantic model.
\item We introduce series base convolutions to SSM to learn multi-scale base semantic representation, and follow a hidden layer to perform complex interactions we called "delay convolution", which performs better than traditional CNN paradigms.
%\item The model we proposed makes structured feature(i.e., word of feed/item and history browser behavior or social network of user) widely and easily use in CTR prediction, which evolve the DNN more compatible for CTR prediction.
\item We conduct extensive experiments on both public and Weibo datasets. Results verify the effectiveness of our proposed SSM. Our code\footnote{Experiment code on two public datasets is available on GitHub: https://github.com/niuchenglei/ssm-dnn\label{code}} is publicly available.
\end{itemize}

%auto-ignore

\section{Relatedwork}

Predicting user responses (such as clicks and conversions, etc.), based on historical behavioral data is critical in industrial applications and is one of the main machine learning tasks in online advertising.
Recommending suitable ads for users not only improves user experience, but also significantly increases a company's revenue.

To deal with the curse of dimensionality in the language model, NNLM\cite{NNLM} proposed using the embedding method to learn the distributed representation of each word, and then using the neural network model to learn the probability function which has a profound impact on subsequent researches. 
%Based on the former research, the Word2Vec method\cite{word2vec} was proposed to learn the distributed representation of words and phrases and their combinations continuously. 
Meanwhile, RNNLM\cite{RNNLM} was proposed to improve existing speech recognition and machine translation systems, and used as a baseline for future researches of advanced language modeling techniques. 
These methods laid a solid foundation for later language models.

To capture the high-dimensional feature interactions, LS-PLM\cite{LS-PLM} and FM\cite{FM} use embedding techniques to process high-dimensional sparse inputs and also design the transformation function for target fitting. 
To further improve the performance of the LS-PLM and FM models, Deep Crossing\cite{DeepCross}, Wide\&Deep Learning\cite{widedeep} and YouTube Recommendation CTR model\cite{youtube} propose a new approach, which uses a complex MLP network instead of the original transformation function. 
FNN\cite{FNN} aims to solving it by imposing factorization machine as embedding initializer.
Moreover, PNN\cite{PNN} adds a product layer after the embedding layer to capture high-level feature interaction information and improve the prediction performance of the PNN model.
Based on the design of the Wide\&Deep framework, DeepFM\cite{deepfm} tried to introduce the FM model as a wide module, aiming to avoid feature engineering. 
%However, these methods will convert the embedding vector into a fixed-length vector, which will result in the loss of the user's original preference information. 
DIN\cite{din} adaptively learns user interests from the advertisement historical behavior data by designing a local activation unit. 
The representation vector varies with different advertisements, which significantly improves the expressive ability of the model.
Considering the influence of the ordering of embedding vectors on the prediction results of the model, CNN-MSS\cite{CNN} proposed the greedy algorithm and random generation method to generate multi-feature sequences in the embedding layer which greatly improved the prediction ability of the model, but the calculation time complexity is extremely high with high-dimensional sparse inputs.

In summary, these researches are mainly accomplished by the techniques of the combination of embedding layer and exploring high-order feature interaction, mainly to reduce the heavy and cumbersome feature engineering work.

%auto-ignore
\section{Structured Semantic Model supported DNN}
Different from the explicit intentions expressed through search queries, advertising and recommendation systems lack user explicit inputs which makes DNN-based models easily fall into local optimum or overfitting.
Hence, it is critical to improve the performance of embeddings.
%Following the principle as FNN\cite{FNN} and CNN-MSS\cite{CNN} our method replaces factorization machine with a structured semantic model.
We introduce a series of base convolution and pooling operators in SSM, and generate multi-scale base semantic representations. Experiments indicate that this method performs better than traditional CNN-based models and DNN-based models.

\subsection{Feature Representation and Word Hashing}
Raw original features of CTR prediction models often consist of two types, categorical such as age=25, gender=male, and numberic such as history\_ctr=0.005. 
Categorical features are normally transformed into high dimensional sparse features via one-hot encoding procedure. 
%Mathmetically speaking, for $\forall$ feature $x$ having $m$ unique ids, there is a trivial embedding mapping $x\rightarrow R^m$. 
Specially, for multi-value features like words and tags, usually represented as ${13,26,98,201}$, one-hot encoding creates a mapping of a vector rather than a single value. 
%That is to say, $X\rightarrow T\in{R^m}$. $m$ denotes the dimensionality of vector $X$, and $T[j]$ denotes the $j-th$ element of $X$ and $T[j]\in\{0,1\}$. $\sum_{1}^{m} T[j]=n$, apparently for single value $n=1$, and for vector $n>1$.
% comment
For example, we have feature set of history\_ctr, age, gender, tags from $78$ tag set encoding as:
%\begin{equation}
\[
\underbrace{[0.052]}_{\text{history\_ctr=0.052}} ~ \underbrace{[0,0,1,\ldots,0]}_{\text{age=20}} ~ \underbrace{[0,1]}_{\text{gender=Female}} ~ \underbrace{[1,0,\ldots,0,1,\ldots,1,0]}_{\text{tags=\{Food,MakeUp,Tourist\}}}
\]
%\end{equation}
Numberic features usually remain unchanged, and could be fed into neural network directly.

Different from english text, text segmentation is a tough task in NLP for chinese text, and a bad word segmentation may led bad performance in the later experiment.
Inspired by DSSM\cite{DSSM}, we directly treat individual character as origin feature inplace of word segmentation. 
For chinese or some other languages not like latin, we break a word into single characters(e.g. p,h,o,t,o) with a given word(e.g. photo), and then represent it using a vector.
% comment
%This method has two benefits. One is dimensionality reduction of words dict and bag-of-words term vectors. 
%The other is, there is no need to worry about new word occurence which is more common in Weibo compared to general NLP scenarios.

Table\ref{table:fea_table} lists all features in our experiment. 
%There is two major types, numberic and categorical. 
There is no combination feature because we rely on DNN to perform deep interactions of original features automatically.

\begin{table}
\caption{Statistics of feature sets used in the display advertising system in Weibo.}
\centering
\resizebox{0.47\textwidth}{!}
{
\label{table:fea_table}
\begin{tabular}{llccc}
\toprule
Category & Feature Group & Dimensionality & Type\\ \toprule
\multirow{3}{7em}{Continuous}
& history\_ctr & 1 & float \\ \cline{2-5}
& hierarchy\_ctr & 1 & float \\ \cline{2-5}
& ... & ... & ... & ...  \\ \midrule

\multirow{4}{7em}{Categorical type(single-value)}
& gender & 2 & one-hot \\ \cline{2-5}
& age & $\sim 80$ & one-hot  \\ \cline{2-5}
& location & $\sim 300$ & one-hot \\ \cline{2-5}
& ... & ... & ... & ...  \\ \midrule

\multirow{5}{7em}{Categorical type(multi-value)}
& user\_tag & $\sim 10^5$ &  multi-hot \\ \cline{2-5}
& cust\_tag & $\sim 10^5$ &  multi-hot \\ \cline{2-5}
& user\_interest & $\sim 10^5$ &  multi-hot  \\ \cline{2-5}
& ad\_word & $\sim 10^6$ &  multi-hot  \\ \cline{2-5}
& ... & ... & ... & ...  \\ \bottomrule
\end{tabular}
}
\label{table_feature_set}
\end{table}

\subsection{Base Model(Wide\&Deep)}
With such two forms of features, categorical and numberic form, and in consideration of the influential structure in display advertising and recommender system, we prefer to use wide\&deep model as our base. It's consists of two components:
% comment

\textbf{Wide Component.} 
The wide component can be explained as a linear model in forms $\widehat{y}=\sigma \left (W^Tx+b \right )$. $\widehat{y}$ denotes prediction and $\sigma$ illustrate a sigmoid function, $x=[x_1,x_2,..,x_n]$ is the vector of features, $W=[w_1,w_2,..,w_n]$ denotes the vector of model parameters and $b$ is the bias.

\textbf{Deep Component.} 
The deep component is a feed-forward neural network consists of multi layers of units using categorical and numberic features. 
Categorical features are sparse and high dimensional generated through one-hot or multi-hot encoding. 
In the deep component, these sparse and high dimensional features are transformed into dense and low dimensional real-valued vectors by embedding layer and pooling layer, generally called embedding vectors. 

%Firstly, the embedding matrix is initialized randomly and then trained to minimize the cross-entropy loss with feedforward neural network. 
%For single-valued features, these low dimensional dense embedding vectors are directly fed into the hidden layers of a neural network. Otherwise, for multi-value features, a pooling operator is usually performed before hidden layers to get fixed length vectors for fully connected networks.
%The most popular pooling layers are sum pooling and average pooling, which apply element-wise sum/average operations to the embedding vectors. The following computation as Eq(\ref{eq:deepcomp}) is performed in the hidden layer, where linear units (ReLUs) is usually chosen as the activation function $f$. 
We apply element-wise sum operations to the embedding vectors. The following computation as Eq(\ref{eq:deepcomp}) is performed in the hidden layer, where linear units (ReLUs) is usually chosen as the activation function $f$.

\begin{tiny}
\begin{equation}
\label{eq:deepcomp}
a^{(l+1)} = f(W^{(l)}a^{(l)}+b^{(l)})
\end{equation}
\end{tiny}%

%All the word type features have both embedding layer and pooling layer.

\textbf{Loss.} 
The objective function of base model is the cross-entropy loss function defined as Eq(\ref{eq:loss}), which $i$ denotes the $i-th$ sample, $y_i\in\{0,1\}$ denotes the true label, $\widehat{y_i}$ is the output of the network after sigmoid layer representing the probability of the $i-th$ sample would be clicked.

{
\begin{tiny}
\begin{equation}
\label{eq:loss}
loss = \sum_{i=0}^{N} {y_ilog(\widehat{y_i}) + (1-y_i)log(1-\widehat{y_i})}
\end{equation}
\end{tiny}
}

% figure about deep model
\subsection{The Structure of Structured Semantic Model}

\begin{figure}[!t]
\centering
\includegraphics[height=5in, width=3.2in, keepaspectratio]{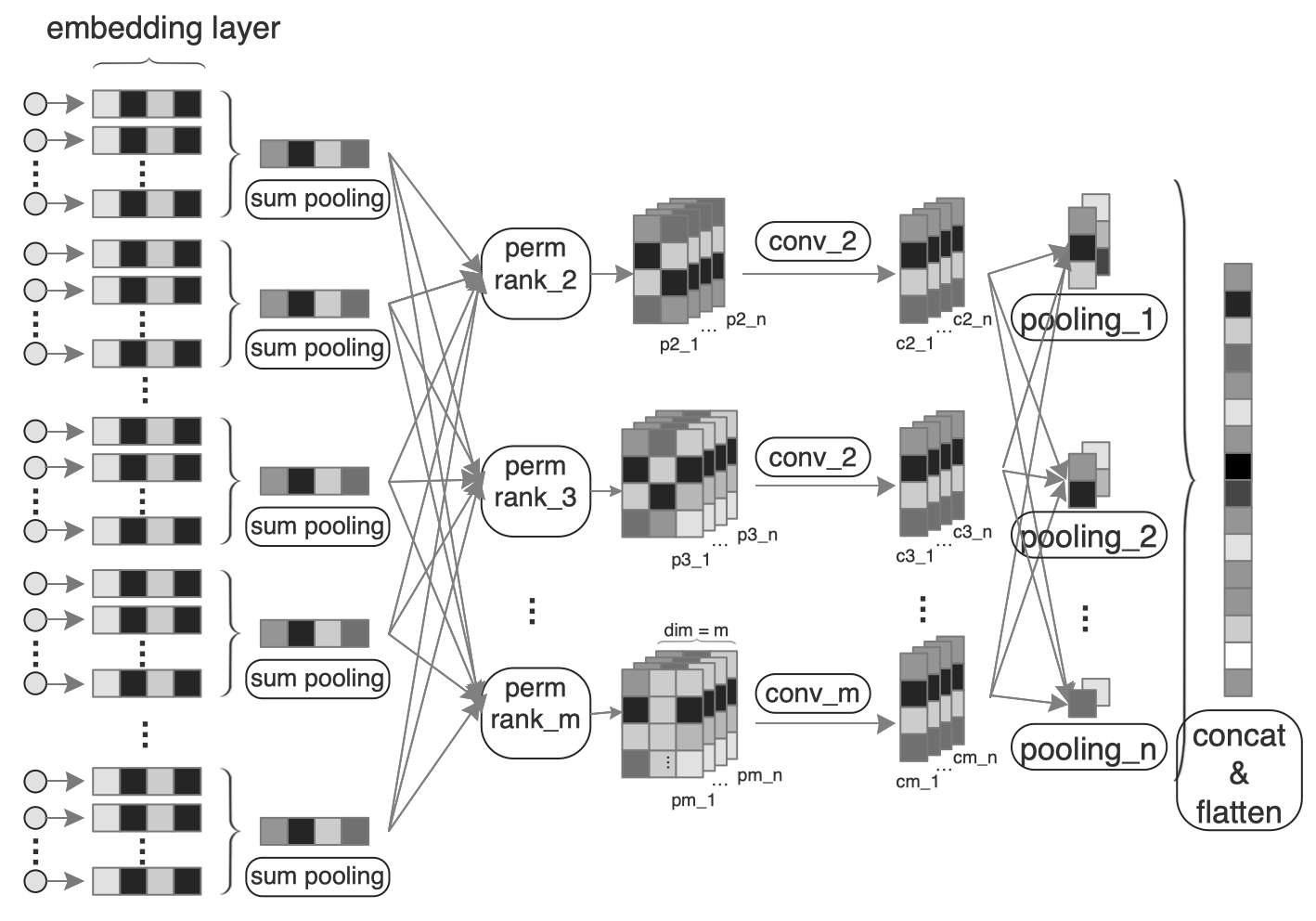}
\caption{The structure of SSM we proposed. It introduces kinds of convolution and pooling operators on embedding vectors, which are able to find structured relations between features.}
\label{model_arch}
\end{figure}

Among all those features of Table\ref{table:fea_table}, the user behavior and ad/item features are used for CTR prediction.
It is difficult to figure out the representations of high level implicit features in feature engineering. 
Hence, it is critical to improve the performance of CTR by figuring out a good representation of high level implicit features.
%The embedding matrix is initialized randomly and then trained with DNN. It is difficult to get a high quality embedding matrix by training the model with end-to-end method due to the vanishing gradient problem. 

DSSM\cite{DSSM} was presented to find the semantic relation between query and doc. Based on DSSM, we introduce an structured convolution pooling network to find the structured semantic relation between high level implicit features. 
The dot-product of two embedding vectors, which is treated as a special convolution and pooling operators as Eq(\ref{eq:pooling}), are commonly used to represent the similarity of two ads/items.

\begin{tiny}
\begin{equation}
\label{eq:pooling}
pooling(Conv({e}_1, {e}_2)) = {e}_1^T{e}_2
\end{equation}
\end{tiny}%

Therefore, we introduce a convolution and pooling layer to figure out the relations between embedding vectors, and optimize the cross-entropy loss with the predicted value ${\widehat{y}}$, and ${\widehat{y}}$ is sigmoid of trainable variables ${W}$ and input vector ${U}$.% as show in Eq(\ref{eq:flatten_lr}).
%{
%\begin{small}
%\begin{eqnarray}
%\label{eq:flatten_lr}
%{\widehat{y}} & = sigmoid({{W}^T{U}} + bias) 
%\end{eqnarray}
%\end{small}
%}

Why convolution and pooling DNN can learn structured semantic relations from the user and item features? 
Our structured semantic model mainly contains three aspects: 

\begin{itemize}
\item i) We introduce an embedding permutation of user and ad/item embedding vectors as Eq(\ref{eq:permutation1}) and $m$ is the rank of permutation to characterize the number of features which interact with each other.

\begin{tiny}
\begin{equation}
\label{eq:permutation1}
C{_n^m} = \frac{{n}!}{{m}!({n-m}!)}
\end{equation}
\end{tiny}%

%\middle/\left.{\frac{\Delta k_i(t+1)}{k_i(t)}}\middle/ \frac{\Delta L(t+1)}{L(t)}=\frac{\Delta k_i(t+1)L(t)}{k_i(t)\Delta L(t+1)}\right.
The space complexity grows exponentially with the rank of feature permutation.
Generally speaking, it is cost prohibitive when the rank is greater than 5.
We permutate user-item combinations instead of all permutations to avoid this problem. The rank less than 4 could get good enough result.
The number of permutation of features is showed in Eq(\ref{eq:permutation}), ${n_1}$ denotes the number of user features and ${n_2}$ denotes the number of ad/item features. 
We defined ${R}$ as a function of ${n_1}$ and ${n_2}$ to produce the combination of feature number, and defined ${P}$ as a cumulative of ${R}$.

\begin{tiny}
\begin{equation}
\label{eq:permutation}
{R}_{n1,n2}^{m} = \sum_{j=1}^{m-1} {C}_{n1}^{j}{C}_{n2}^{m-j}\ \ \ m = 2,3,\ldots n
\end{equation}
\end{tiny}%
\begin{tiny}
\begin{equation}
\label{eq:permutation}
{P}_{n1,n2}^{r} = \sum_{i=2}^{r} {R}_{n1,n2}^{i}\ \ \ r = 2,3,\ldots n
\end{equation}
\end{tiny}

\item ii) 1-d convolution is defined as Eq(\ref{eq:conv1d}), ${e}$ denotes the embedding vector and ${K}$ denotes a convolution kernel. We adapt many kinds of 1-d linear convolution operators to describe how embedding vectors interact with each other and treat dot-product as a special convolution operator as same as other linear kernel functions(ie. [-1,1], [1,1]).
%comment  as same as
Normally, the size of convolution kernel is set from [1,2] to [1,4] coresponding to the rank of permutation.

\begin{tiny}
\begin{equation}
\label{eq:conv1d}
{Conv\_1d}([{k}_1,{k}_2],{e}_1, {e}_2) = {k}_1{e}_1 + {k}_2{e}_2
\end{equation}
\end{tiny}

The 1-d convolution kernel shown in \ref{eq:kernel} was used in our practice (permutation rank 3 was used), which describes 5 linear and 1 special convolution kernels, and they are all orthogonal to each other.
In addition, traditional CNN-based models train multi trainable convolution kernels(like [$w_1$,$w_2$]) also could get good enough result by lots of training epochs, but fixed orthogonal linear convolution kernels followed by a hidden layer could have much better performance, and we will explain it in later chapters.

{
\begin{tiny}
\begin{eqnarray}
\label{eq:kernel}
%\begin{matrix} 0 & 1 \\ 1 & 0 \end{matrix}\quad
%\begin{pmatrix} 0 & -i \\ i & 0 \end{pmatrix}\\
\begin{bmatrix} {dot\ product} \end{bmatrix}\quad
\begin{bmatrix} 1 & -1 \\ 1 & 1 \end{bmatrix}\quad
\begin{bmatrix} -1 & 1 & 1 \\ 1 & -1 & 1 \\ 1 & 1 & -1 \end{bmatrix}\quad
%\begin{Bmatrix} 1 & 0 \\ 0 & -1 \end{Bmatrix}\\
%\begin{vmatrix} a & b \\ c & d \end{vmatrix}\quad
%\begin{Vmatrix} i & 0 \\ 0 & -i \end{Vmatrix}
\end{eqnarray}
\end{tiny}
}

%Convolutional Neural Networks based Click-Through Rate Prediction with Multiple Feature Sequences
CNN-MSS\cite{CNN} pointed out that using convolutional networks with multiple feature sequences could produce much more non-linear and deep representations, but training variables also get multiplied by the number of sequences which may not converge or could fall into local optimum easily.
As show in our practice, use SSM instead of CNN-MSS could converge into better results.

\item iii) We introduce a multi-scale pooling layer to figure out the relations between vectors.
Pooling operations behind embedding vectors, can be viewed as a scale function similar to pooling in image and voice fields.
Finally, we apply the logistic function \cite{lr} to the concat and flatten layer, and we name the last flatten layer of SSM multi-scale basic semantic representation of user and ad/item features.
\end{itemize}

\paragraph{\textbf{Put It All Together}} the flatten layer of SSM can be denoted as:
\begin{tiny}
\begin{eqnarray}
\label{eq:flatten}
{P} & = Perm({r}, \langle Emb({\theta_1},f_1),\ Emb({\theta_2},f_2),\ldots,\ Emb({\theta_n},f_n)\rangle) \\
{U} & = Flatten(Pooling({s}, Conv({k}, {P})) 
\end{eqnarray}
\end{tiny}

As shown in Eq(\ref{eq:flatten}), ${\theta_1}$ and ${\theta_2}$ denote the embedding matrix of user and item feature, $f_1$/$f_2$ denotes the input features, and ${r}$ is the rank of permutation which corresponds to the shape of convolution kernel ${k}$. 
The convolution kernels are matrices\ref{eq:kernel}, ${s}$ is the parameter of pooling operator(3,11 and 19 was set in our practice).
%We can find that SSM perform semantic mining by convolution and pooling on embedding combinations, which pass the gradient of trainable variables in very short way.

In our practice, firstly, we embed 3 user features and 2 item/ad features with the shape of embedding vector ${e}$ set to $[5K,100]$ and 20 sequences with shape $[100,2]$ and $[100,3]$ generated by a permutation of rank 2 and 3.
Secondly, convolute these 20 sequences with 6 kernels whose shapes are $[1,2]$ and $[1,3]$ to generate 70 vectors .
Thirdly, take pooling operation over 80 vectors with 3 kinds of pooling operators with window size set to 3, 7 and 13 to get 3 matrices ${p1_{[98,70]}}$, ${p2_{[32,70]}}$ and ${p3_{[15,70]}}$.
Lastly, flatten the 3 pooling matrices ${p1,p2,p3}$ to a flattened vector ${Flt_{[1,10150]}}$ as the input of LR.

With the approach metioned above, the SSM learns feature embeddings and its multi-scale basic semantic representation for DNN which improves AUC of CTR prediction a lot.
The embedding learned by SSM can describe the semantic information over features efficiently.
%and the last flatten layer of SSM describe multi-scale base semantic information of features.

\subsection{Multi-Scale Base Semantic Representation}
We named the last flatten layer of SSM as Multi-Scale Base Semantic Representation, which together with the embedding vectors are fed into DNN as inputs. 

\begin{figure}[!t]
\centering
\includegraphics[height=4in, width=2.3in, keepaspectratio]{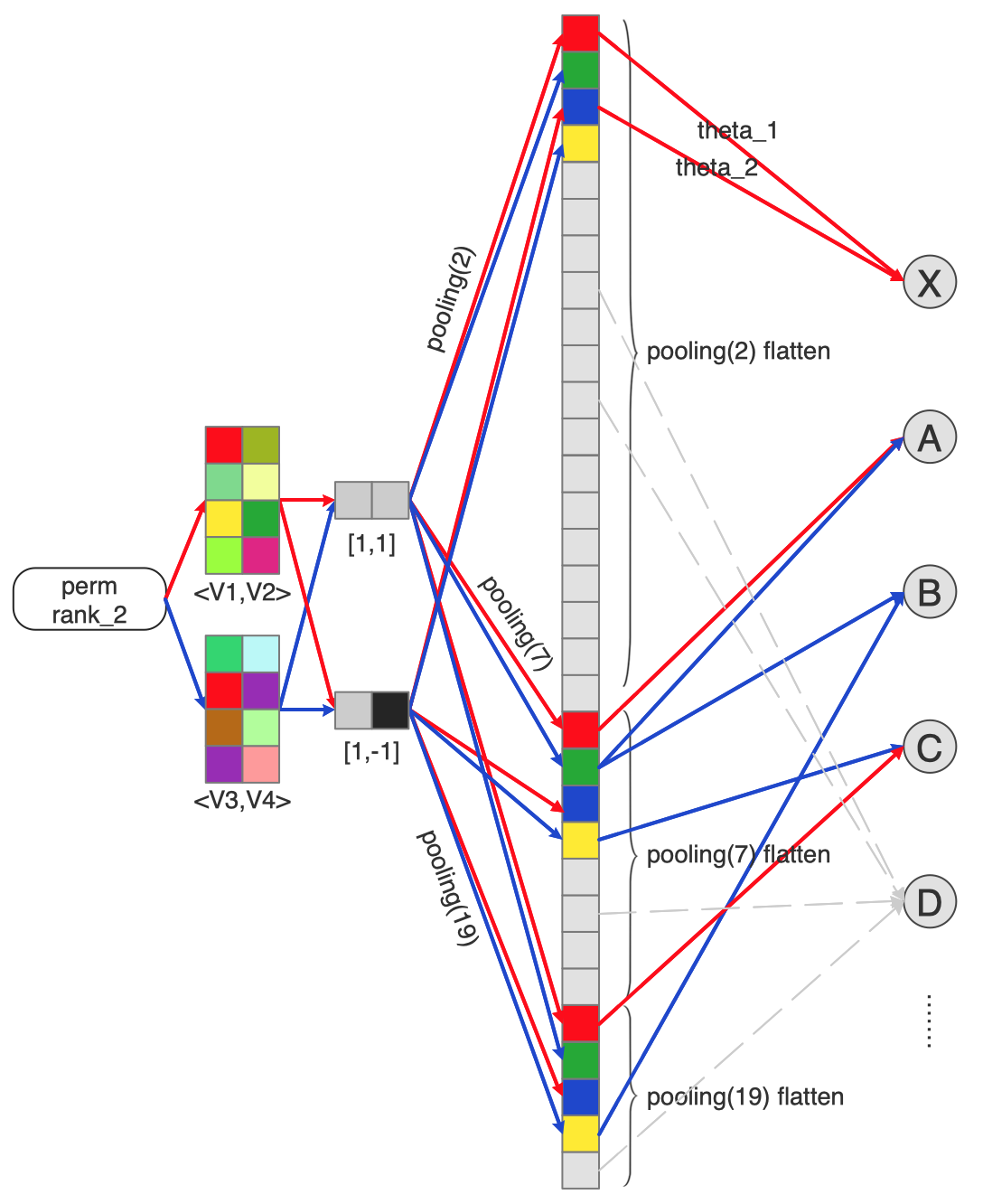}
\caption{Multi-Scale Base Semantic Representation with hidden layer. The multi-scale base semantic representation followed by a hidden layer constructs non-linear relations between embedding vectors. The node 'X' illustrates a convolution kernel $[\theta_1+\theta_2, \theta_1-\theta_2]$ over $[V_1, V_2]$,  node 'B' illustrates the relations between $[V_3, V_4]$ in different scales. All nodes are fully connected with multi-scale base semantic representation just like node 'D'. }
\label{model_base_scale}
\vspace{-0.4cm}
\end{figure}

\textbf{Base Semantic} For signal processing, wavelet and fourier transformation play a critical role to transform signal from time domain to frequency domain. The base functions of wavelet are sine and cosine function. 
We propose the base function to CTR prediction, which contains two types of linear convolution kernels with shapes of $[1,2]$ and $[1,3]$, and a special convolution kernel with dot-product.
%, and thus Multi-Scale Base Semantic Representation with base functions can reveal various representions in semantic domain. 
In CNN-MSS\cite{CNN} and other convolution related approaches, the convolution kernels are usually trainable, and a large number of feature mapping in CNN might results in overfitting and computational consumption. 
On the contrary, our base-formed kernels(\ref{eq:kernel}) are fixed and not trainable. Linear relations are represented by using properties of base functions and non-linear relations are represented by a hidden layer. 
%need fix

\textbf{Multi-Scale.}
The multi-scale can be demonstrated by multiple pooling operators with various size.
% need fix

In traditional CNN-based models, the convolution and pooling operators are independent. 
Pooling operator always follows convolution operators which means training process first determines the function, and then determines what kind of scale to put it on. 

Aside from base functions mentioned above, there also exist lots of interactions among different pooling segments within the multi-scale embedding vector, such as interactions between word and word vectors, word and classification vectors, classification and classification vectors. We define it as multi-scale interaction modeling problem.

That process produces $Num(conv)*Num(pooling)$ parameters to train. 
For NLP and CTR prediction, interaction affections exist between multi-scale embeddings and the number of parameters of traditional CNN-based models will be too huge to get a good result.

% In detail, the complex relations between features are decomposed into two parts as following:

% \begin{itemize}
% \item i) Firstly, using click or not as label to supervise learn the embeddings with SSM, and the last flatten layer of SSM actually representes multi-scale relations over base functions. 
% \item ii) Secondly, by follow a hidden layer after SSM, we generate various non-linear or complex interactions which is a linear combination of base semantic representations .
% \end{itemize}

% interaction affections 指特征交叉影响
\textbf{Delay Convolution.} The SSM we proposed in this paper can solve the problems efficiently. 
First of all, we produce series of base semantic representation using base functions (the matrices\ref{eq:kernel} mentioned above), and then use hidden layer to choose convolution kernel types and pooling scale at the same time, which is named is {\bfseries delay convolution}. 

Node of hidden layer can be defined as Eq(\ref{eq:hidden_node}). 
SSM followed by a hidden layer gets better performance than traditional CNN-based models. Training convolution kernels and scale together reduces computation load and improves convergence speed.
Additionally, the nodes of hidden layer in Fig\ref{model_base_scale} can not only describe combinations across different base functions but also different scales which brings more non-linear relations. For instance, the relations between words and classifications can be expressed as a linear combination of two pooling operators with different scales. 
\begin{tiny}
\begin{equation}
\label{eq:hidden_node}
{N_k} = Relu \left ( \sum_{i=0}^{N} \theta_{i} Pooling({S}, Conv({K}, {P})) \right )
\end{equation}
\end{tiny}

\subsection{SSM supported Wide\&Deep}
\begin{figure}[!t]
\centering
\includegraphics[height=7in, width=3.2in, keepaspectratio]{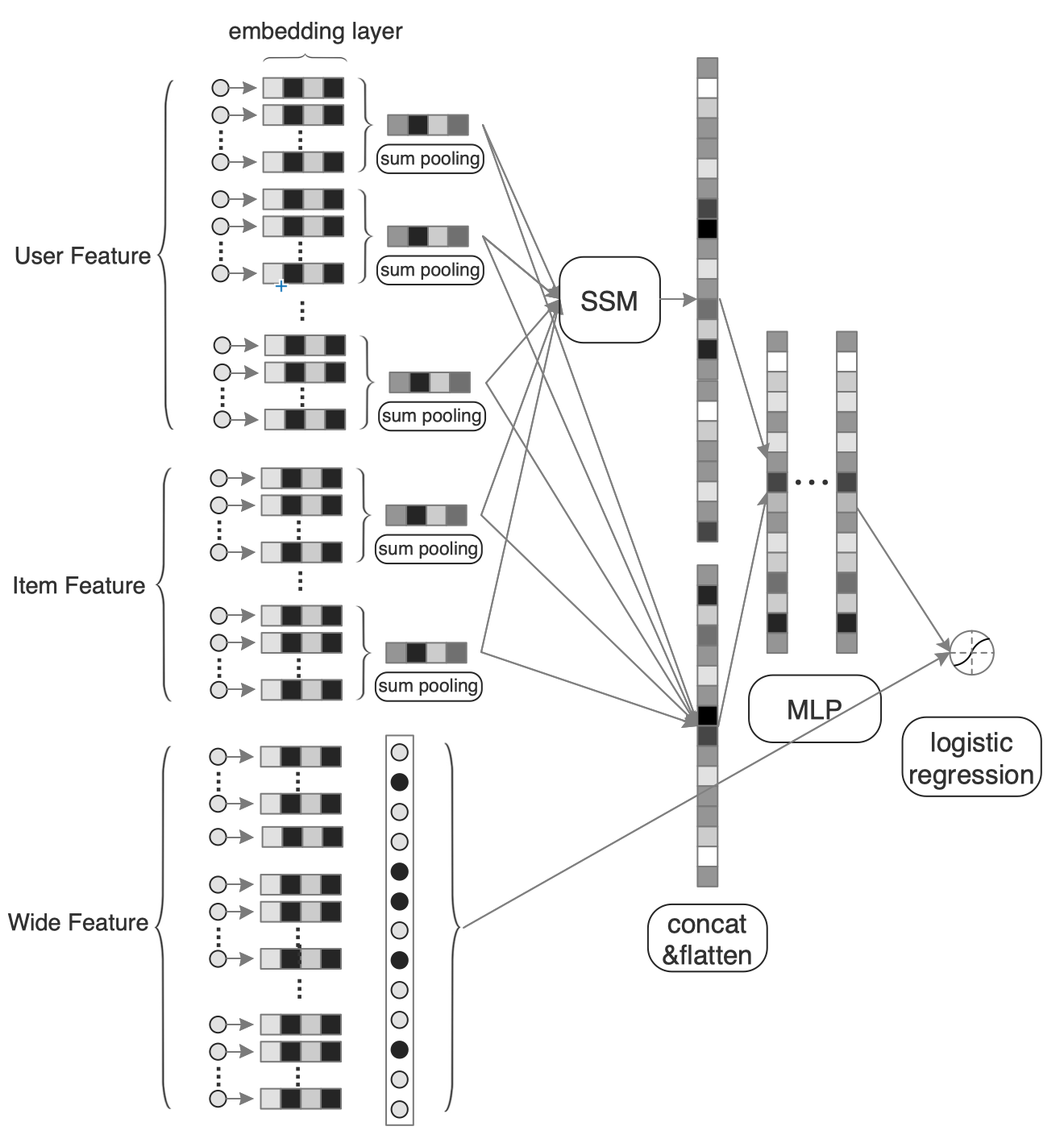}
\caption{SSM supported Wide\&Deep Architecture. Above show the structure of SSM supported Wide\&Deep we proposed, it concat the output of SSM as partial input of DNN as same as the embeddings.}
\label{model_whole}
\vspace{-0.4cm}
\end{figure}

The multi-scale base semantic representation learned by SSM associates with inputs in DNN by simply concated to the flatten embeddings.
In Eq(\ref{eq:logits}), ${\widehat{y}}$ denotes the prediction value of CTR which is the output of sigmoid function $\sigma$ and $W_{wide}$ denotes the weights of origin features $X$ and feature engineering features $\phi (X)$, $W_{deep}$ denotes the weights of the output of DNN, and the red arrowhead denotes the multi-scale base semantic representation concat to the input of DNN.
In Eq(\ref{eq:logits_af}), $a^{(l_0)}$ denotes the input of DNN which is a concatation of embeddings ${e_i}$ and the last layer of SSM ${U}$, and then $a^{(l_0)}$ follows a standard MLP transform in Eq(\ref{eq:deepcomp}).
The whole architecture of Wide\&Deep-SSM show as Fig\ref{model_whole}.

\begin{tiny}
\begin{equation}
\label{eq:logits}
{\widehat{y}} = \sigma \left (W_{wide}^T{[X,\phi (X)] + W_{deep}^T{a^{(l_f)}}} \right )
\end{equation}
\end{tiny}
\begin{tiny}
\begin{equation}
\label{eq:logits_af}
{a^{l_0}} = concat\_flatten \left ( {e}_1, {e}_2, \ldots, {e}_n, {U} \right )
\end{equation}
\end{tiny}

%\subsection{Implementation Details}

%\subsection{Deep Interest Network}
%\label{din}

%auto-ignore

\section{Experiments} \label{exp}

In this section, our experiments are discussed in detail including the datasets, experimental settings, model comparisons and the corresponding analysis. 

Both the public datasets and experiment codes are made available on github\textsuperscript{\ref{code}}.
%\footnote{Experiment codes on two public datasets are available at GitHub: https://github.com/zhougr1993/DeepInterestNetwork}.

\begin{figure*}[!t]
\centering
\includegraphics[height=7.2in, width=6.2in, keepaspectratio]{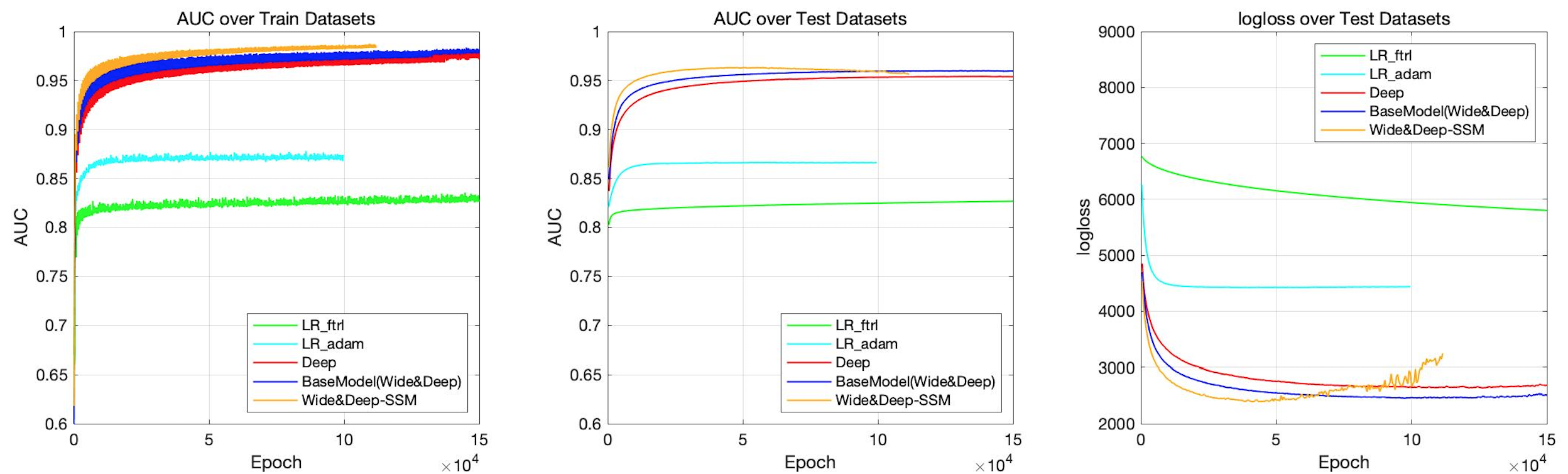}
\caption{Performances of different model on Weibo Dataset. Wide\&Deep-SSM shows the improvement over test logloss and AUC.}
\label{logloss_auc}
%\vspace{-0.4cm}
\end{figure*}

\begin{table*}[!t]
\caption{Model Comparison on Avazu, MovieLens and Weibo Datasets. All the lines calculate RelaImpr by comparing with BaseModel on each dataset respectively.}
\centering
\begin{threeparttable}
\begin{tabular}{lccccccc}
\toprule
\multirow{2}{*}{Model} & \multicolumn{2}{c}{Avazu(Electro).}  & \multicolumn{2}{c}{MovieLens.}  & \multicolumn{2}{c}{Weibo.} \\ 
& AUC & RelaImpr & AUC & RelaImpr & AUC & RelaImpr \\ \midrule
LR  & 0.6525 & -24.0\% & 0.6984 & -25.2\% & 0.8371 & -12.58\%  \\
Deep  & 0.6963 & -2.19\% & 0.7576 & -2.87\% & 0.8742 & -2.98\%  \\
%\textbf{Deep-SSM} &\textbf{0.XXXX} & \textbf{-X.XX}\% &\textbf{0.XXXX} &\textbf{ X.XX}\% &\textbf{0.XXXX} & \textbf{-X.XX}\% &\textbf{0.XXXX} & \textbf{-X.XX}\%\\
BaseModel(Wide\&Deep) & 0.7007 & $\sim$ & 0.7652 & $\sim$ & 0.8857 & $\sim$ \\
\textbf{Wide\&Deep-SSM} &\textbf{0.7011} & \textbf{0.2}\% &\textbf{0.7754} &\textbf{3.9}\% &\textbf{0.9109} & \textbf{6.55}\%  \\
\bottomrule
\end{tabular}
%\begin{tablenotes}%[para,flushleft]
%    \item[\sim] Means no different with self.
%\end{tablenotes}
\end{threeparttable}
\label{table:exptable_public}
\end{table*}
%\vspace*{-0.4cm}

\subsection{Datasets and Settings}
%\textbf{Amazon Dataset\footnote{http://jmcauley.ucsd.edu/data/amazon/}} 
We conduct experiments on Weibo dataset and two opensource datasets. 

\textbf{Avazu Dataset\footnote{https://www.kaggle.com/c/avazu-ctr-prediction}.} 
In Avazu dataset, which is provided in 2014 Kaggle competition. The dataset contains 40 million samples with 22 fields for ten consecutive days, and all fields are used in our experiments. 
We use the samples of first nine days as training set and the samples of the tenth day for evaluation.

\textbf{MovieLens Dataset\footnote{https://grouplens.org/datasets/movielens/20m/}.} 
MovieLens data\cite{movie_lens} contains 138,493 users, 27,278 movies, 21 categories and 20,000,263 samples. 
To make it suitable for CTR prediction task, we transform it into a binary classification data. 
Original user rating of the movies is continuous value ranging from 0 to 5. We label the samples as positive if the rating is above 3(e.g. 3.5,4.0,4.5,5.0), otherwise is negative. 
%Among all 20,000,264 samples(movie.csv), of which $80\%$ samples are randomly selected into training set (about 16,197,321 samples) and the rest $20\%$ into the test set (about 3,802,943 samples). The task is to predict whether user will rate a given movie is positive based on historical behaviors. Features include movie\_id, movie\_cate\_id and user rated history movie\_id\_list, history movie\_cate\_id\_list.

\textbf{Weibo Dataset.}
In Weibo dataset, we collected impression logs from the online display advertising system, of which two weeks' samples are used for training and samples of the following day for testing. 
The size of training and testing set is about 1 billions and 0.1 billion respectively. 31 fields of features are separated into three categories (user profile, ad information and context information) for each sample. 
%Due to the size of weibo dataset, we set the batch size to be 10000 and use Adam as the optimizer and set learning rate at 0.0001.

\subsection{Model Comparison}
\textbf{LR.}\cite{lr} Logistic regression (LR) is a widely used in CTR prediction task because training is fast and results can be easily explained. We treate it as a weak baseline.

\textbf{BaseModel(Wide\&Deep Model).}\cite{widedeep} The google's wide\&deep model has been widely used in real industrial applications. We treat it as the benchmark model. 
It consists of two parts: wide model, which handles the manually designed cross product features and deep model, which automatically extracts non-linear relations among features. We set it as BaseModel. 
%Wide\&Deep needs expertise feature engineering on the input of the "wide" module. We take cross-product of user behaviors and ad information as wide inputs.

\textbf{Wide\&Deep-SSM}. Defferent from Wide\&Deep, the result of SSM concated to the input of DNN.

\subsection{Performance of the Experiment}
AUC is one of the most popular evaluation metrics for CTR prediction which measures the goodness of order by ranking all the ads with predicted CTR, including intra-user and inter-user orders. 
%To make the experiment more reliable, we adopt RelaImpr introduced in \cite{cauc} to measure relative improvement over models. 
We adopt RelaImpr introduced in \cite{cauc} to measure relative improvement over models
%For a random guesser, the value of AUC is $0.5$. 
It is defined as follows:

\begin{small}
\begin{equation}
RelaImpr = \left(\frac{\text{AUC(measured model)} -0.5}{\text{AUC(base model)} - 0.5} - 1\right) \times 100\%.
\end{equation}
\end{small}

\begin{figure}[!t]
\centering
\includegraphics[height=6.2in, width=3.2in, keepaspectratio]{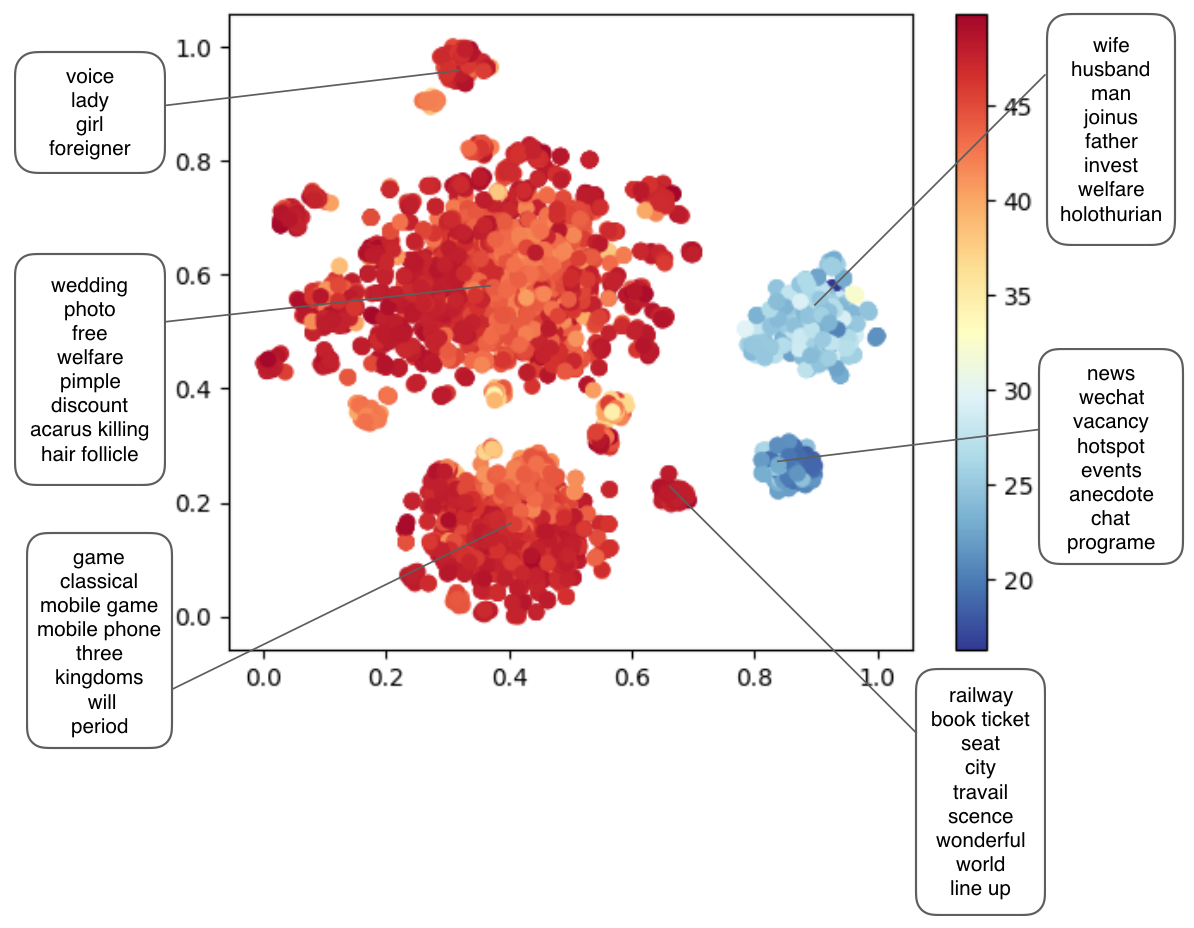}
\caption{Visualization of flatten layer of samples in SSM. Shape of points represents a single sample. Color of points corresponds to CTR prediction value.}
\label{diagram}
\end{figure}

Fig\ref{logloss_auc} shows the training and testing AUC on LR, base model and Wide\&Deep-SSM. 
The statistics of all the above datasets is shown in Table\ref{table:exptable_public}.

\par From Table\ref{table:exptable_public}, we can see that, in terms of RelaImpr, all deep networks perform significantly better than LR, and Wide\&Deep-SSM has the best performance. 

\subsection{Visualization of SSM}
We take samples of different tags to visualize the learned multi-scale base semantic representation by two-dimensional scatter plot using t-SNE\cite{tsne}. 
The cluster points represent the same types of items/ad texts. 
That means words from similar item/ad are almost aggregated together, which demonstrates the aggregation ability of SSM flatten vector. 
Besides, it's also a Heat Map that different intensity indicate the prediction values (high intensity reveals high click prediction). 
Fig\ref{diagram} shows the distribution of users' inclination to click advertisements in semantic space. 
As we can see, SSM can express a preference distribution in semantic space efficiently.

%auto-ignore

\section{Conclusions}
In this paper, we focus on the task of CTR prediction in online advertising with rich structured data.
The performance of the embeddings of those structured data learned by DNN-based model is a bottleneck for the performance of the CTR prediction.
To improve the performance of the embeddings, a novel approach named SSM is designed to pre-train the embeddings by structured semantic model to get semantic relations between features, and fine tune it with DNN variables.
Additionally, we introduce series base convolution instead kinds of trainable convolutions to learn the multi-scale base semantic representation, and follow a hidden layer for complex interactions which we called delay convolution.
SSM get good performance both in opensource dataset or Weibo dataset, and can be extend to other DNN-based models easily.

\clearpage

\bibliographystyle{named}
\bibliography{structured_semantic_model_supported_deep_neural_network.bib}

\end{document}